# Resource-Conscious Modeling for Next-Day Discharge Prediction Using Clinical Notes

Ha Na CHO[a,1], Sairam SUTARI[a], Alexander LOPEZ[b], Hansen BOW[b], Kai ZHENG[a]

[a] *University of California, Irvine, Donald Bren School of Information, Department of Informatics*
[b] *University of California, Irvine, Department of Neurosurgery*
ORCiD ID: Ha Na CHO https://orcid.org/0000-0001-8033-6644

**Abstract.** Timely discharge prediction is essential for optimizing bed turnover and resource allocation in elective spine surgery units. This study evaluates the feasibility of lightweight, fine-tuned large language models (LLMs) and traditional text-based models for predicting next-day discharge using postoperative clinical notes. We compared 13 models, including TF-IDF with XGBoost and LGBM, and compact LLMs (DistilGPT-2, Bio_ClinicalBERT) fine-tuned via LoRA. TF-IDF with LGBM achieved the best balance, with an F1-score of 0.47 for the discharge class, a recall of 0.51, and the highest AUC-ROC (0.80). While LoRA improved recall in DistilGPT2, overall transformer-based and generative models underperformed. These findings suggest interpretable, resource-efficient models may outperform compact LLMs in real-world, imbalanced clinical prediction tasks.

**Keywords.** Hospital discharge planning, fine-tuning transformer models, clinical notes, natural language processing, discharge prediction, low-resource AI

## 1. Introduction

Accurate and timely hospital discharge prediction is critical for optimizing hospital operations [1], especially in elective spine surgery, where bed turnover and recovery must be closely managed [2]. While next-day discharge forecasting can support proactive care coordination, it remains underutilized in practice due to the subjectivity of decision-making and limited integration of narrative clinical data.

Prior work has largely focused on structured data models trained on demographics, vitals, procedures [3], often overlooking the rich contextual signals embedded in clinical notes. Recent advances in clinical natural language processing (NLP) have explored transformer-based models, such as BERT, and GPT, for unstructured text modeling [4],[5], but these approaches assume access to high-resource infrastructure and offer limited insight into deployment feasibility. Although recent work [6] has applied large language models (LLMs) for general postoperative risk prediction, their work focused on broader surgical cohorts and relied on GPU-based systems.

---

[1] Corresponding Author: Author Name, Contact details.

This study addresses these gaps by benchmarking diverse modeling paradigms, ranging from TF-IDF with traditional machine learning to fine-tuned LLMs, for next-day discharge prediction in elective spine surgery. We emphasize performance-efficiency tradeoffs and practical, lightweight approaches suitable for deployment in resource-constrained clinical settings.

**2. Methods**

*2.1. Study Design, Cohort, and Data Source*

We conducted a comparative modeling study to predict next-day discharge for patients undergoing elective spine surgery using real-world clinical notes. The dataset comprised 2,000 de-identified post-operative notes from electronic health record (EHR) at the University of California, Irvine Health, covering patients aged 18 years and older who underwent spine surgery between 2018 and 2023. Each note was labeled with a binary outcome indicating whether the patient was discharged the following day. This study was approved by the Institutional Review Board of the University of California, Irvine (IRB #4537).

*2.2. Preprocessing*

All notes underwent Unicode normalization, ASCII decoding, and removal of non-standard characters. Text was lowercased, punctuation removed, and whitespace normalized. Clinical abbreviations were expanded and discharge-related terms were masked via rule-based matching for machine learning features only. Notes rendered empty after masking were replaced with 'EMPTY'. TF-IDF representations were extracted using SpaCy's lemmatization pipeline. Sentence embeddings were obtained via all-MiniLM-L6-v2 (MiniLM) and Bio_ClinicalBERT (BCB). Long sequences were chunked (10% stride) to fit models' limits (512 for BERT, 1024 for GPT).

*2.3. Modeling Approaches*

All models were trained and evaluated on stratified train, validation, test splits (64/16/20), preserving class distribution and ensuring consistency across experiments. Model evaluations used three random seeds (42, 123, 999) for robustness. TF-IDF features were extracted from preprocessed clinical notes using unigrams and bigrams (max_features=5,000). Three classifiers were trained: Logistic Regression (LR), XGBoost (XGB), and LightGBM (LGBM) (Figure 1). Class weighting was applied for imbalance correction, with gradient boosting models using sample weighting. Hyperparameters were optimized via five-fold stratified RandomizedSearchCV.

Sentence embeddings were generated using MiniLM (384D) and BCB. Notes were split into sentences, encoded, and pooled via mean aggregation. The same classifiers and training protocols as the TF-IDF setup were applied, including hyperparameter tuning. Moreover, we fine-tuned two lightweight transformer models for next-day discharge prediction: DistilGPT2 and BCB. DistilGPT2 was fine-tuned as a causal language model using HuggingFace's trainer API with batch size 8 for 2 epochs, learning rate 5e-5, weight decay 0.01, and warmup ratio 0.06. Input sequences were chunked or truncated to fit the 1024-token limit. For BCB, we fine-tuned using class-weighted cross-entropy loss (weights: 1.0, 5.0 for majority and minority classes), with AdamW optimization and fp16 mixed precision. A 10% stratified sample was used and the training ran for 1 epoch with batch size 4. For all models, we selected the thresholds that maximized the F1-score on the precision-recall curve.

We defined best performance primarily using the F1-score for the discharge class (class 1), as accurate identification of patients eligible for next-day discharge is the primary objective of this study. This metric balances both precision (avoiding false positives that could risk premature discharge) and recall (ensuring eligible patients are not missed). While we also report AUC-ROC, accuracy, and per-class precision/recall, the F1-score offers a clinically grounded, threshold-sensitive measure of practical utility. For a more comprehensive view, we supplement this with AUC-ROC and class 0 (non-discharge) performance to ensure model safety and overall balance. All experiments were conducted on an AWS EC2 instance using Python 3.11.

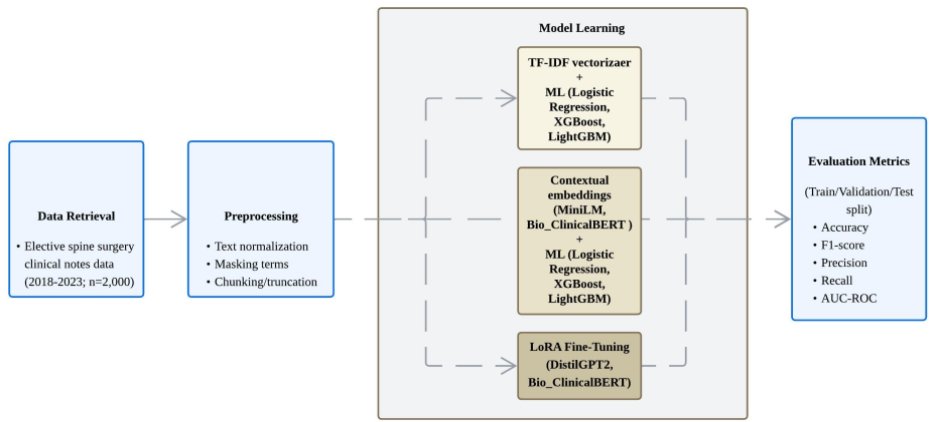

**Figure 1.** Pipeline for lightweight clinical NLP model evaluation.

## 3. Results

We evaluated modeling performance across three modeling strategies: TF-IDF with ML classifiers, embedding-based models using MiniLM and BCB, and fine-tuned generative LLMs. Table 1 presents evaluation results across 13 models for next-day discharge prediction using precision, recall, and F1-scores (per class), and AUC-ROC. TF-IDF with LR achieved the highest F1-score for class 1 (0.48), with recall of 0.71 and AUC-ROC of 0.79, indicating strong discharge case detection. TF-IDF with LGBM followed closely with F1-score of 0.47 and the highest AUC-ROC (0.80), making it the most balanced model overall. In contrast, TF-IDF with XGB demonstrated the highest accuracy (0.82) and class 0 F1-score (0.89) but performed poorly for class 1 (F1-score 0.38), indicating potential bias toward predicting non-discharge.

**Table 1.** Performance of models for next-day discharge prediction (T=optimized threshold; 1=discharge; 0=non-discharge). Includes TF-IDF+ML, transformer embeddings, and fine-tuned LLMs.

| Model | Accuracy | Precision | | Recall | | F1-Score | | AUC-ROC |
|---|---|---|---|---|---|---|---|---|
| | | 0 | 1 | 0 | 1 | 0 | 1 | |
| TF-IDF + LR | 0.73 | 0.92 | 0.36 | 0.74 | 0.71 | 0.82 | **0.48** | 0.79 |
| TF-IDF + XGB | 0.82 | 0.87 | 0.45 | 0.92 | 0.33 | 0.89 | 0.38 | 0.78 |
| TF-IDF + LGBM | 0.81 | 0.90 | 0.44 | 0.87 | 0.51 | 0.88 | 0.47 | **0.80** |
| MiniLM + LR | 0.59 | 0.88 | 0.24 | 0.59 | 0.62 | 0.70 | 0.34 | 0.63 |
| MiniLM + XGB | 0.82 | 0.83 | 0.16 | 1.00 | 0.01 | 0.90 | 0.01 | 0.62 |
| MiniLM + LGBM | 0.74 | 0.84 | 0.21 | 0.85 | 0.20 | 0.84 | 0.20 | 0.60 |
| BCB + LR | 0.82 | 0.83 | 0.17 | 0.99 | 0.01 | 0.90 | 0.01 | 0.65 |
| BCB + XGB | 0.82 | 0.58 | 0.33 | 0.98 | 0.04 | 0.90 | 0.07 | 0.60 |
| BCB + LGBM | 0.83 | 0.83 | 0.44 | 1.00 | 0.01 | 0.90 | 0.03 | 0.63 |
| DistilGPT2 (T=0.15) | 0.45 | 0.89 | 0.21 | 0.39 | 0.76 | 0.54 | 0.32 | 0.59 |
| DistilGPT2+LoRA (T=0.42) | 0.27 | 0.89 | 0.20 | 0.36 | 0.79 | 0.51 | 0.43 | 0.59 |
| BCB (T=0.42) | 0.22 | 0.91 | 0.18 | 0.06 | 0.97 | 0.12 | 0.30 | 0.56 |
| BCB + LoRA (T=0.32) | 0.25 | 0.90 | 0.18 | 0.10 | 0.94 | 0.18 | 0.30 | 0.55 |

Transformer-based embeddings (MiniLM and BCB) underperformed in discharge class detection. MiniLM with LR and MiniLM with LGBM achieved only moderate AUC-ROC (0.60-0.63), with class 1 F1-scores maxing at 0.34. MiniLM with XGB failed to detect discharge cases. BCB variants achieved high accuracy and class 0 recall (>0.99) but very low class 1 F1-scores (<0.03), indicating severe class imbalance issues. BCB with XGB showed slight improvements but still lacked generalizability.

Interestingly, LoRA-based fine-tuning modestly improved generative model performance. DistilGPT2 with LoRA achieved an F1-score of 0.43 for class 1 and recall of 0.79, a notable improvement over base DistilGPT2 (F1-score=0.32). However, AUC-ROC remained moderate (0.59), and LoRA did not enhance BCB performance, possibly due to domain or task misalignment. BCB with LoRA exhibited signs of overfitting, with perfect precision but poor recall for class 1.

These findings highlight that traditional text representations combined with machine learning models can outperform more complex transformer and generative architectures, particularly when the outcome variable is highly imbalanced. Furthermore, while accuracy was often highest in models heavily favoring the majority class, macro-averaged metrics and per-class scores reveal that such models fail to identify discharges effectively. Hence, F1-score for class 1, paired with AUC-ROC, offers a more meaningful metric of real-world utility in this clinical classification context.

## 4. Discussion and Conclusions

This study benchmarked diverse modeling strategies for next-day discharge prediction from clinical notes. Traditional models, particularly TF-IDF combined with logistic regression or LightGBM, consistently outperformed fine-tuned transformers and contextual embeddings in detecting minority discharge cases. These results reinforce the utility of traditional NLP techniques, particularly in low-resource settings. Despite modest F1-score ceilings, TF-IDF models demonstrated strong balance. These results align with a recent work [3] demonstrating strong performance in discharge forecasting using EHR-based XGBoost models, and compact transformers can rival larger models with reduced computational cost [7]. A recent survey also highlights the growing relevance of small language models in healthcare, offering efficiency and adaptability, and potential for deployment in limited clinical settings [8].

Contextual sentence embeddings from MiniLM and BCB struggled to capture discharge signals, likely due to overfitting on majority class patterns and insufficient task alignment. This aligns with prior findings that BERT-based models can be disproportionately influenced by features associated with the majority class [9]. LoRA-based fine-tuning modestly enhanced recall in DistilGPT2, suggesting potential for improving sensitivity in low-resource generative models. However, performance gains were inconsistent across architectures, offering no benefit for BCB, possibly due to model complexity or domain mismatch. Recent study also showed that TF-IDF representations can outperform transformer-based embeddings [10]. These findings collectively demonstrate that accurate interpretable discharge forecasting is achievable without reliance on large-scale LLMs or GPUs. TF-IDF models and compact lightweight alternatives can offer robust performance and practical utility in low-resource clinical environments.

Nonetheless, several limitations must be noted. We used a single-institution dataset and focus on elective spine surgeries, potentially restricting generalizability. Future work should explore user-facing explainability and integration of structured EHR data. In conclusion, traditional text-based models remain highly competitive for clinical prediction tasks under resource constraints. With careful tuning, they can surpass more complex architectures in both effectiveness and interpretability, offering a scalable pathway for decision support in discharge planning.